\documentclass{article}
\usepackage[nonatbib,preprint]{nips_2018}
\usepackage{algorithm}
\usepackage{algorithmicx}
\usepackage{algpseudocode}
\usepackage{float}
\usepackage{bbm}

\usepackage{microtype}
\usepackage{graphicx}
\usepackage{subfigure}
\usepackage{booktabs} % for professional tables
\usepackage{amsmath}

\usepackage{amssymb}
\usepackage{amsfonts}

\newtheorem{definition}{Definition}
\newtheorem{theorem}{Theorem}
\newtheorem{corollary}{Corollary}

\newtheorem{lemma}{Lemma}
\usepackage{hyperref}

\DeclareMathOperator*{\argmax}{arg\,max}

\newcommand{\tr}{\mathrm{tr}}
\newcommand{\bfw}{\mathbf{w}}
\newcommand{\bfx}{\mathbf{x}}
\title{Faster Convergence \& Generalization in DNNs}
\author{
Gaurav Singh \\
Department of Computer Science\\
University College London,\
UK \\
\texttt{gaurav.singh.15@ucl.ac.uk} \\
\And
John Shawe-Taylor \\
Department of Computer Science\\
University College London,\
UK \\
\texttt{j.shawe-taylor@ucl.ac.uk } \\
}
\PassOptionsToPackage{noend}{algpseudocode}% comment out if want end's to show
\usepackage{algpseudocode}% http://ctan.org/pkg/algorithmicx

\errorcontextlines\maxdimen

\makeatletter
\newcommand*{\algrule}[1][\algorithmicindent]{\makebox[#1][l]{\hspace*{.5em}\vrule height .75\baselineskip depth .25\baselineskip}}%

\newcount\ALG@printindent@tempcnta
\def\ALG@printindent{%
    \ifnum \theALG@nested>0% is there anything to print
        \ifx\ALG@text\ALG@x@notext% is this an end group without any text?
            \addvspace{-3pt}% FUDGE for cases where no text is shown, to make the rules line up
        \else
            \unskip
            \ALG@printindent@tempcnta=1
            \loop
                \algrule[\csname ALG@ind@\the\ALG@printindent@tempcnta\endcsname]%
                \advance \ALG@printindent@tempcnta 1
            \ifnum \ALG@printindent@tempcnta<\numexpr\theALG@nested+1\relax% can't do <=, so add one to RHS and use < instead
            \repeat
        \fi
    \fi
    }%
\usepackage{etoolbox}
\patchcmd{\ALG@doentity}{\noindent\hskip\ALG@tlm}{\ALG@printindent}{}{\errmessage{failed to patch}}
\makeatother

\begin{document}
% \twocolumn[
% \icmltitle{Generalization \& Faster Convergence in Deep Networks}
% \icmlsetsymbol{equal}{*}
% \begin{icmlauthorlist}
% \icmlauthor{Gaurav Singh}{ucl}
% \icmlauthor{John Shawe-Taylor}{ucl}

% \end{icmlauthorlist}

% \icmlaffiliation{ucl}{University College London, London, United Kingdom}

% \icmlcorrespondingauthor{Gaurav Singh}{g.singh.15@ucl.ac.uk}
% \icmlcorrespondingauthor{John Shawe-Taylor}{j.shawe-taylor@ucl.ac.uk}

% % You may provide any keywords that you 
% % find helpful for describing your paper; these are used to populate 
% % the "keywords" metadata in the PDF but will not be shown in the document
% \icmlkeywords{Deep Learning, Fast Convergence, Generalization in Deep Network}

% \vskip 0.3in
% ]
\maketitle
\begin{abstract}
Deep neural networks have gained tremendous popularity in last few years. They have been applied for the task of classification in almost every domain. Despite the success, deep networks can be incredibly slow to train for even moderate sized models on sufficiently large datasets. Additionally, these networks require large amounts of data to generalize. The importance of speeding up convergence and generalization in deep networks can not be overstated.  In this work, we develop an optimization algorithm based on generalized-optimal updates derived from minibatches that lead to faster convergence. 
% We show that the learned network is robust to adversarial noise and over-fitting. 
Towards the end, we demonstrate on two benchmark datasets that the proposed method achieves two orders of magnitude speed up over traditional back-propagation and is more robust to noise/over-fitting.
\end{abstract}
\section{Introduction}
Deep networks have gained immense popularity in the last few years. These networks were originally theorized in the 90's, but did not become immediately popular due to the lack computational power. Advancements in gpu technology over the years has led to these neural networks getting recognition. We have moved from using traditional multi-layer perceptrons (MLP) to more sophisticated networks like convolutional neural networks, recurrent neural networks, long shot-term memory networks etc. 

These deep models are highly expressive and require large amounts of training data to avoid over-fitting. In addition, the  training time required for most of these networks remains significant. Plus, there have been no theoretically sound ways to generalize these networks, except for using empirical methods like dropout and parameter weight regularization. Most theoretical analysis in the area begins by making assumptions on the structure of the network. Importance of the ability to train these networks in less time with smaller training data with generalization can not be overstated.

One possible direction for speeding up the training involves speeding up the computations involved in traditional back-propagation \cite{mathieu2013fast, vasilache2014fast}. It can improve the speed of computation but does not reduce the number of epochs required for convergence. Such methods that improve on the computational efficiency of back-propagation do not lead to better generalization in the learned model. Also, the improvement in speed are generally not orders of magnitude higher over traditional optimizers.

Another direction to achieve speed up would be to improve upon traditional back-propagation i.e. using sgd or its variants. It has been shown that sgd consists of two phases: drift phase and diffusion phase \cite{shwartz2017opening}. It is well known that sgd is slow to converge and the reason for slow convergence is attributed to the diffusion phase, where learning rate needs to be lowered to avoid overshooting, leading to slow convergence. It is argued that the algorithm tries to compress the network during diffusion phase, and therefore does not focus on loss reduction. 
% Also, it is not straightforward to obtain estimates on the speed of convergence or generalization using sgd.  

One relevant work by Shawe-Taylor \emph{et al.} \cite{SHAWETAYLOR1990575} tried to approximate a neural network optimization process with a linear program. The linear program restricted the output nodes from moving in the wrong direction after update. It  made a reasonable assumption that if the updates were small the second order effects could be ignored even for a network with non-linear activations. But, the work did not make any efforts towards generalizing the updates. In this work, we aim to develop an approach that generates optimal and generalized updates that can lead to faster convergence with improved generalization of the model. The proposed method leads the model in the optimal direction to significantly decrease loss with each update.

Our contribution in this work can be summarized briefly as: 1) we develop an algorithm that computes generalized optimal updates that lead to faster convergence; 2) the proposed algorithm achieves two orders of magnitude speed up over traditional back-propagation on two benchmark datasets; 3) we bound the classification error of the network after k iterations; 4) we demonstrate that the learned network is more robust to adversarial noise and over-fitting.

\section{Related Works}
Deep neural networks started to gain popularity in late 90's \cite{hochreiter1997long, lecun1998gradient, lawrence1997face}. These networks existed before in theory but the lack of computational power required to train such network rendered them useless for most practical purposes. Advancement in gpu technology enabled orders of magnitude faster training for one particular type of deep networks, referred to as convolutional deep networks \cite{lawrence1997face}. Consequently, these networks have been used to obtain state-of-the-art performances in image/text/video/audio classification tasks \cite{krizhevsky2012imagenet, zeiler2014visualizing, simonyan2014very}.

Further attempts to speed up computations of the convolution function in these networks have been made \cite{mathieu2013fast, vasilache2014fast}. One direction for making these deep convolution networks fast would involve speeding up of computations \cite{mathieu2013fast, denton2014exploiting} using techniques like fourier transforms. Another direction would be to approximate the computations exploiting the redundancies arising due to linear structure in these networks, instead of explicitly computing all floating point operations in the convolution layers \cite{denton2014exploiting}.

Despite these attempts, the training process is still painfully slow, owing to the slow speed of convergence obtained using traditional back-propagation. The algorithm used often to train these networks is stochastic gradient descent. SGD has its merits, it is simple to implement; and  acts like an implicit regularizer \cite{zhang2016understanding} by leading to solutions with smaller norm. But, it is slow to converge, as has been explained in Shwartz-Ziv \emph{et al.} \cite{shwartz2017opening}. It shows that convergence through SGD has two phases: drift phase and diffusion phase. In the drift phase it explores the multidimensional space of solutions. When it begins converging, it arrives at the diffusion phase where it is extremely chaotic and the convergence rate slows to a crawl. It is argued that the network learns to compress during this phase. 

Attempts to achieve faster convergence using a modification of the optimization algorithm have been made in the past with little success. Wilamowski \emph{et al.} \cite{wilamowski2001algorithm} used a modification of Levenberg-Marquardt algorithm (that relies on the second order derivatives) to get more optimal updates. One work that tried to fix the problems discussed above, and is nearest to this work,  was done by Shawe-Taylor \emph{et al.} \cite{SHAWETAYLOR1990575}. It formulates neural network optimization as a linear program. It solves the LP to compute the most optimal update that leads to a movement in the correct direction for all output nodes simultaneously. Another work that solved very similar optimization problem was done by Hochreiter \emph{et al.} \cite{hochreiter2005optimal}. They applied the same optimization technique albeit to speed up and improve memory in RNNs. But, they did not draw theoretical justifications to ensure the update was generalized and excluded experiments on deep CNNs.

On the issue of generalization, there have been works to give theoretical bounds/measures. But, most such works make strong assumptions on the structure of the network. One such early work assumed that the network is linear and consists of 3 layers \cite{nakajima2005generalization}. It considers an empirical bayes approach where some of the parameters are regarded as hyper-parameters. Other works made assumptions on the nature of regularizer \cite{xie2015generalization} to obtain desired bounds on generalization.

\section{Method}
\subsection{Generalisation from a mini-batch}
\label{sec:gen}
The approach we adopt to choosing our update direction is motivated by an application of a learning theory bound that generalises from the mini-batch to the rest of the training and test examples. We choose a direction that improves the outputs on the majority of the mini-batch in terms of the hinge loss for each output label. Given that the mini-batch is a randomly chosen sample, and that the selected direction satisfies good bounds on generalisation, we infer that the update will not only improve the hinge loss of the mini-batch but also with high probability the hinge loss of the majority of the rest of the training examples and a large fraction of the underlying distribution generating the test data.

Our analysis only applies if we ignore second order terms:
$
f^j(\bfw + \delta\bfw, \bfx) \approx f^j(\bfw, \bfx) + \langle \nabla f^j(\bfw,\bfx),\delta \bfw \rangle,
$
where $f^j(\bfw,\bfx)$ is the $j$-th output of the network with weights assigned values $\bfw$ and input $\bfx$.
Given the use of RELU activation functions, provided we are not at an inflexion point of the activation of any of the neurons for any of the inputs, there will be an open set around the current weight vector in which this equation will hold exactly. By controlling the norm of the update vector, we minimise the effect of the second order terms.

Hence, our task is to find an update direction $\delta \bfw$ that satisfies the following optimisation for a mini batch of size $\ell$ and a problem with $K$ classes:

\begin{algorithmic}
	\State minimise $\frac{1}{2}\|\delta \bfw\|^2 + C \sum_{i=1}^\ell\sum_{j=1}^K \xi_{ij}$
	\State Subject to: $y_{ij}\langle \delta \bfw_i, \nabla f^j(\bfw, \bfx_i)\rangle \geq \epsilon - \xi_{ij}$
	\State $\xi_{ij} \geq 0$, $i = 1, \ldots, \ell; j = 1, \ldots, K$
\end{algorithmic}

where $\epsilon$ is our target reduction in hinge loss for each example $i$ and output class $j$. Note that $y_{ij}$ indicates which class example $i$ belongs to, controlling whether this class's output should be increased or reduced. Here we have introduced a parameter $C$ to trade off the norm of the update vector with the overall amount by which points in the mini-batch fail to reduce their hinge loss by $\epsilon$.

This optimisation corresponds exactly to a support vector machine with $\ell K$ examples but with the target output reduced from $1$ to $\epsilon$, and with the threshold fixed at $0$. The dual of this optimisation is given by

\begin{algorithm}
	\caption{Dual Update Calculation}
	\label{dual}
\begin{algorithmic}
	\State max $\epsilon\sum_{ij}\alpha_{ij} - \frac{1}{2}\sum_{ijkl} \alpha_{ij}y_{ij}\alpha_{kl}y_{kl}\kappa((\bfx_i, j), (\bfx_k,l))$
	\State subject to: $C\geq \alpha_{ij} \geq 0$,	
\end{algorithmic}
\end{algorithm}
where $\kappa$ is the linear kernel taking inner products between the gradient vectors for the corresponding inputs/outputs, that is
\[
\kappa((\bfx_i, j), (\bfx_k,l)) = \left\langle \nabla f^j(\bfw,\bfx_i), \nabla f^l(\bfw,\bfx_k) \right\rangle
\]
This also means that we can apply standard margin based generalisation bounds to infer the expected reduction in hinge loss of samples outside the mini-batch, given the observed reduction in the mini-batch and the observed margin. We now introduce the relevant theoretical bounds.

\begin{definition}[Rademacher complexity]
	\index{Rademacher complexity}For a sample $S=\left\{ \mathbf{x}_{1},\ldots ,%
	\mathbf{x}_{\ell }\right\} $ generated by a distribution $\mathcal{D}$ on a
	set $X$ and a real-valued function class $\mathcal{F}$ with domain $X$, the
	\emph{empirical Rademacher complexity} of $\mathcal{F}$ is the random
	variable
	\begin{equation}
	\hat{R}_{\ell }(\mathcal{F})=\mathbb{E}_{\mathbf{\sigma }}\left[ \left.
	\sup_{f\in \mathcal{F}}\left\vert \frac{2}{\ell }\sum_{i=1}^{\ell }\sigma
	_{i}f\left( \mathbf{x}_{i}\right) \right\vert _{{}}\right\vert \mathbf{x}%
	_{1},\ldots ,\mathbf{x}_{\ell }\right] \text{,}
	\end{equation}%
	where $\mathbf{\sigma }=\left\{ \sigma _{1},\ldots ,\sigma _{\ell }\right\} $
	are independent uniform $\left\{ \pm 1\right\} $-valued (Rademacher) random
	variables. 
% 	The \emph{Rademacher complexity }of $\mathcal{F}$ is%
% 	\begin{equation}
% 	R_{\ell }(\mathcal{F})=\mathbb{E}_{S}\left[ \hat{R}_{\ell }(\mathcal{F})%
% 	\right] =\mathbb{E}_{S\mathbf{\sigma }}\left[ \sup_{f\in \mathcal{F}%
% 	}\left\vert \frac{2}{\ell }\sum_{i=1}^{\ell }\sigma _{i}f\left( \mathbf{x}%
% 	_{i}\right) \right\vert \right] .
% 	\end{equation}
\end{definition}

\begin{theorem}
	\label{radebound}Fix $\delta \in \left( 0,1\right) $ and let $\mathcal{F}$
	be a class of functions mapping from $Z$ to $[0,1]$. Let $(\mathbf{z}%
	_{i})_{i=1}^{\ell }$ be drawn independently according to a probability
	distribution $\mathcal{D}$. Then with probability at least $1-\delta $ over
	random draws of samples of size $\ell $, every $f\in \mathcal{F}$ satisfies
	\begin{eqnarray*}
		\mathbb{E}_{\mathcal{D}}\left[ f(\mathbf{z})\right] &\leq &\hat{\mathbb{E}}%
		\left[ f(\mathbf{z})\right] +R_{\ell }(\mathcal{F})+\sqrt{\frac{\ln
				(2/\delta )}{2\ell }} \\
		&\leq &\hat{\mathbb{E}}\left[ f(\mathbf{z})\right] +\hat{R}_{\ell }(\mathcal{%
			F})+3\sqrt{\frac{\ln (2/\delta )}{2\ell }}\text{.}
	\end{eqnarray*}
\end{theorem}

\begin{definition}
	\index{slack variable}%
	\index{margin}%
	\index{functional margin}For a function $g:X\rightarrow \mathbb{R}$, we
	define its \emph{margin} on an example $(\mathbf{x},y)$ to be $yg(\mathbf{x}%
	) $. The \emph{functional margin} of a training set $S=\left\{ (\mathbf{x}%
	_{1},y_{1}),\ldots ,(\mathbf{x}_{\ell },y_{\ell })\right\} $, is defined to
	be%
	\begin{equation}
	m(S,g)=\min_{1\leq i\leq \ell }y_{i}g(\mathbf{x}_{i})%
	\text{.}
	\end{equation}%
	Given a function $g$ and a desired margin $\gamma $ we denote by $\xi
	_{i}=\xi \left( (\mathbf{x}_{i},y_{i}),\gamma ,g\right) $ the amount by
	which the function $g$ fails to achieve margin $\gamma $ for the example $(%
	\mathbf{x}_{i},y_{i})$. This is also known as the example's \emph{slack
		variable}%
	\begin{equation}
	\xi _{i}=\left( \gamma -y_{i}g(\mathbf{x}_{i})\right) _{+}\text{,}
	\end{equation}%
	where $\left( x\right) _{+}=x$ if $x\geq 0$ and $0$ otherwise.
\end{definition}

\begin{theorem} \cite{Shawe04}
	\label{margthm}Fix $\gamma >0$ and let $\mathcal{F}$ be the class of
	functions mapping from $Z=X\times Y$ to $\mathbb{R}$ given by $f\left(
	\mathbf{x},y\right) =-yg(\mathbf{x})$, where $g$ is a linear function in a
	kernel-defined feature space with norm at most $B$. Let
	\begin{equation}
	S=\left\{ (\mathbf{x}_{1},y_{1}),\ldots ,(\mathbf{x}_{\ell },y_{\ell
	})\right\}
	\end{equation}%
	be drawn independently according to a probability distribution $\mathcal{D}$
	and fix $\delta \in \left( 0,1\right) $. Then with probability at least $%
	1-\delta $ over samples of size $\ell $ we have%
	\begin{eqnarray*}
	\mathbb{E}_{\mathcal{D}}\left[ \max(1,(\gamma -yg(\mathbf{x}))_+)\right] &\leq&
	\frac{1}{\ell \gamma }\sum_{i=1}^{\ell }\xi _{i}+\frac{4B}{\ell \gamma }\sqrt{%
		\tr(\mathbf{K})}\\
	&&+3\sqrt{\frac{\ln (2/\delta )}{2\ell }}\text{,}
	\end{eqnarray*}%
	where $\mathbf{K}$ is the kernel matrix for the training set and $\xi
	_{i}=\xi \left( (\mathbf{x}_{i},y_{i}),\gamma ,g\right) $.
\end{theorem}
The theorem indicates conditions under which we can expect the reductions in hinge loss we have secured on the mini-batch will generalize to hinge loss reductions across the training and test sets. 
% Hence, we propose to translate these results into an algorithm that solves Algorithm~\ref{dual} with mini-batch sizes that guarantee generalizatio. A sequence of these updates will reduce hinge loss across training and test sets eventually resulting in both low training and test error. 
\textbf{Note}, the bounds depend on the norm of the svm (B) learned using data and are data dependant.  

\begin{lemma}
    \label{lemma}
	Let $\varepsilon(\gamma, \delta) = \mathbb{E}_{\mathcal{D}}\left[ \max(1,(\gamma -yg(\mathbf{x}))_+)\right]$ (as defined in Theorem \ref{margthm})  and $H = \mathbb{E}_{\mathcal{D}}\left[ \ell(f(x), y)\right]$ where $H$ is the true hinge loss of the network.
	 Then  after $(i+1)_{th}$ iteration we have with probability at least $1-\delta/k $:
	
	\begin{equation}
	    H_{i+1} \leq H_i - \left( \bigtriangleup_i - \varepsilon_i(\gamma_i, \delta/k) \right) + h_i
	\end{equation}
	
	where $\bigtriangleup$ is the step size for the update (as defined in Algorithm \ref{algo}) and $h_i$ are the second order effects that we assume to be negligible.
\end{lemma}

\begin{theorem}
\label{maintheorem}
    Let $n$ be the number of output nodes that have $\ell(.)>0$ and 
    \begin{equation}
	S=\left\{ (\mathbf{x}_{1},y_{1}),\ldots ,(\mathbf{x}_{\ell },y_{\ell
	})\right\}
    \end{equation}
    be drawn independently according to a probability distribution $\mathcal{B}$ such that $\mathcal{B}_{(x_i, y_i)} \propto n_{i}*\mathcal{D}_{(x_i, y_i)}$. Then applying Lemma \ref{lemma} $k-$times and taking union over $\delta$ we have with probability at least $1-\delta$:
    \begin{equation}
	    H_{k} \leq H_0 - \sum_{i=1}^{k} \left( \bigtriangleup_i - \varepsilon_i(\gamma_i, \delta/k) \right) + \sum_{i=1}^{k}h_i
	\end{equation}
    where $H_0$ is the initial hinge loss of the network, and $H_{k}$ is the loss after $k$ iterations.
\end{theorem}

\begin{corollary}
 Let $\hat{y} = \argmax_j f^j(\bfx, \bfw)$ and $e = \mathbbm{1}(\hat{y}\neq y)$. Then $e=1$ if and only if $\sum_{j}\ell(f^j(\bfx, \bfw),y^j)\geq 2$. We have with probability at least $1-\delta$:
 \begin{equation}
    E_k \leq \frac{1}{2} H_{k}     
 \end{equation}
 where $E_k = \mathbb{E}_\mathcal{D} \left[e\right]$ is the classification error of the network after k iterations.
\end{corollary}

\subsection{Optimization}
We use the updates derived in Section \ref{sec:gen} to optimize the deep network. We compute gradient of each output node with respect to the parameters of the network given an instance $x$. We collect the gradients corresponding to all instances $x$ in the minibatch. We feed these gradients as input to a linear svm. The gradients of positive output nodes are labelled as $+1$ and gradients of negative output nodes are labelled as $-1$. The svm returns the update ($\delta w$) which is then applied to the current parameters of the deep network after being multiplied by step size (see Algorithm 2). Note that it is not essential to consider every  negative gradient corresponding to all classes for a given input. 

The value of $B\sqrt{tr(\mathbf{K})}/\ell$ (defined in Theorem 2) can be computed for each minibatch. If the generalization obtained is not satisfactory then the svm batch size is doubled and svm is further regularized. Once the sample size is sufficiently large it would ensure that each update derived from the minibatch reduces the error over both train and test set. We do not apply updates that have a loose bound as they could diverge the optimization away from the solution.  
% The loss decreases by a certain amount over both train and test sets with each iteration. 

\algrenewcommand{\algorithmicrequire}{\textbf{Input:}}
\algrenewcommand{\algorithmicensure}{\textbf{Output:}}
\algrenewcommand{\algorithmicforall}{\textbf{for each}}
\newcommand{\To}{\textbf{to}\xspace}
% Not stated in manual, \Return and \algorithmicreturn are defined, but no \algorithmicstate, why?
\newcommand{\Dosth}{\State \textbf{Do Something}\xspace}
% Note no space before \xspace
\newcommand{\Please}[1]{\State \textbf{#1}}
\newcommand{\ForEach}{\ForAll}

\begin{algorithm}[]
    \caption{\textsc{Training}}
     \begin{algorithmic}
     % here
     \Function{trainMinibatch}{$\mathcal{D}$, $\mathcal{C}$, $r$}
     \State $\mathcal{G} \gets \{\}$
     \For {$\forall  (x,y) \in \mathcal{D}$}
        \For{$o \in [1, 2 \cdots |\mathcal{C}|-1, |\mathcal{C}|]$}
        % \State //Obtain gradient of hinge loss on output 
        % \State //node $o$ wrt parameters in the network
            \State \# obtain gradient of hinge loss on output node $o$
            \State $g_{o}$ $\gets$ getGradient($x, y_{o}$) 
            \If{$y_{o}= 1$}
                \State $\mathcal{G}$ $\gets$ $\mathcal{G}\cup \{(g_{o}, +1)\}$ 
            \Else
                \State $\mathcal{G}$ $\gets$ $\mathcal{G}\cup \{(g_{o}, -1)\}$ 
            \EndIf
            \State $\delta w$ $\gets$ trainSVM($\mathcal{G}$, $r$)
            \State \Return $\delta w$
            
        \EndFor    
     \EndFor
     
     \EndFunction
     
     \Function{train}{$\mathcal{D}$, batch\_size}
     \State step\_size $= 0.1$; $r =1.0$
     \For{$i \in [1, 2,  \cdots, num\_iter]$}
        \State $\mathcal{D}_s$ $\gets$ generateMinibatches($\mathcal{D}$, batch\_size)
        \State $\delta w$ $\gets$ trainMinibatch($\mathcal{B}_s$, $\mathcal{C}$, $r$)
        \State bound\_term = $B\sqrt{tr(\mathbf{K})}/|\mathcal{D}_s|$
        \If{bound\_term $>$ threshold}
            \State batch\_size $\gets$ 2 * batch\_size
            \State \# regularizer for svm (lower value $\rightarrow$ more regularization)
            \State $r \gets   0.1*r$ 
        \EndIf
        \State $w$ $\gets$ $w + step\_size*\delta w$ 
     \EndFor
     \State \Return $w$
     \EndFunction
         \label{algo} 
     \end{algorithmic}
    
\end{algorithm}

\subsection{Cost comparison of updates}
Currently, the proposed method can take more run time than SGD for convergence due to our crude implementation. Below we discuss technical solutions that can speedup our updates.
\begin{itemize}
    \item Sample-wise gradients: We require gradients for each sample in the proposed algorithm. It is not efficient to obtain such gradients in existing libraries like tensorflow\cite{Abadi:2016:TSL:3026877.3026899} as these were designed to minimize a scalar loss.  Currently, we obtain these gradients one at a time using a batch size of one. A library can be designed that provides unaggregated gradients and save save time due to concurrent computations and in-bulk copying of data to the gpu. 
    \item Computing kernel gram matrix: We can compute the kernel gram matrix required for solving the svm  dual formulation in the gpu\cite{fatahalian2004understanding}. The kernel gram matrix corresponds to a minimatch and has a reasonable size that can fit in the gpu. 
    \item Solving svm: A lot of research has been done on efficiently solving an svm. Accelerated optimization algorithms are available \cite{sun2015accelerating}. More recently, there have been some works to solve an svm using a gpu using parallel computations \cite{cotter2011gpu}. 
\end{itemize}
These technical improvements were beyond the scope of this work. Hence, we restrict the discussion to comparing epochs required for convergence instead of run time. 
\section{Experimentation}
In this section, we first briefly describe the different datasets used for experimentation. Afterwards, we describe the experimental settings, followed by a brief section on hyper-parameter tuning. Finally, we present the obtained results on two benchmark datasets.

% % \begin{comment}
% % \begin{figure}[ht!]
%     \centering
%     \includegraphics[scale=0.30]{images/ratio.png}
%     \caption{We plot the ratio of decrease in train hinge loss to the decrease in minibatch hinge loss  $\frac{\bigtriangleup \ell_{train}}{\bigtriangleup\ell_{batch}}$ as training progresses. We can see that the updates derived using our method are generalized since it translated into decrease in loss over the entire train set.}
%     \label{fig:ratio}
% \end{figure}
% \end{comment}

\subsection{Dataset}
We test our method on two bench-marking image datasets: CIFAR-10 \cite{krizhevsky2009learning} and MNIST \cite{lecun2016jc}.

\begin{itemize}
    \item CIFAR-10 \cite{krizhevsky2009learning}: It consists of 60000 32x32 colour images in 10 classes, with 6000 images per class. The images have 3 channels, namely RGB, therefore have a depth of 3. It contains 10 classes e.g. cat, deer, dog, automobile etc., and classes are completely mutually exclusive. 
    \item MNIST \cite{lecun2016jc}: It consists of images of handwritten digits in binary. It has a training set of 60,000 examples, and a test set of 10,000 examples. The images are binary and each pixel value is either 0 or 1. The classes are again mutually exclusive from each other. 
\end{itemize}
\begin{figure}
    \centering
    \includegraphics[scale=0.30]{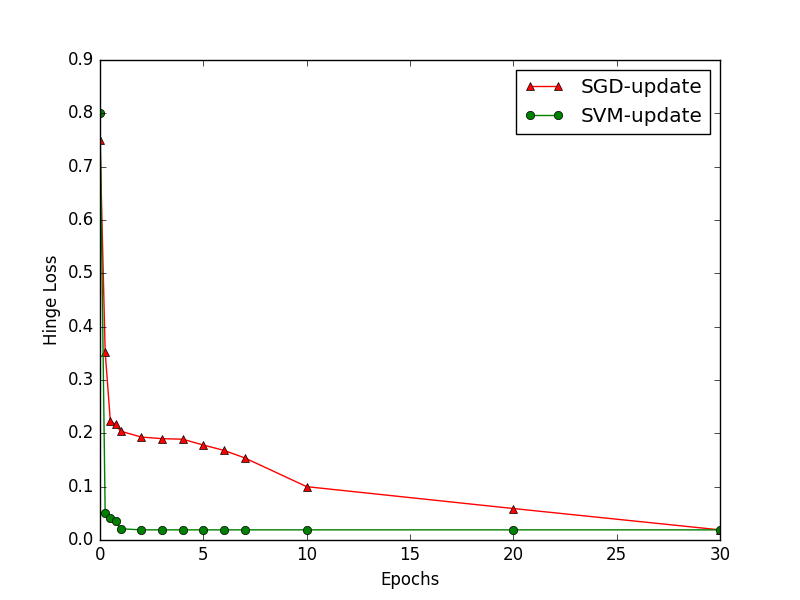}
    \includegraphics[scale=0.30]{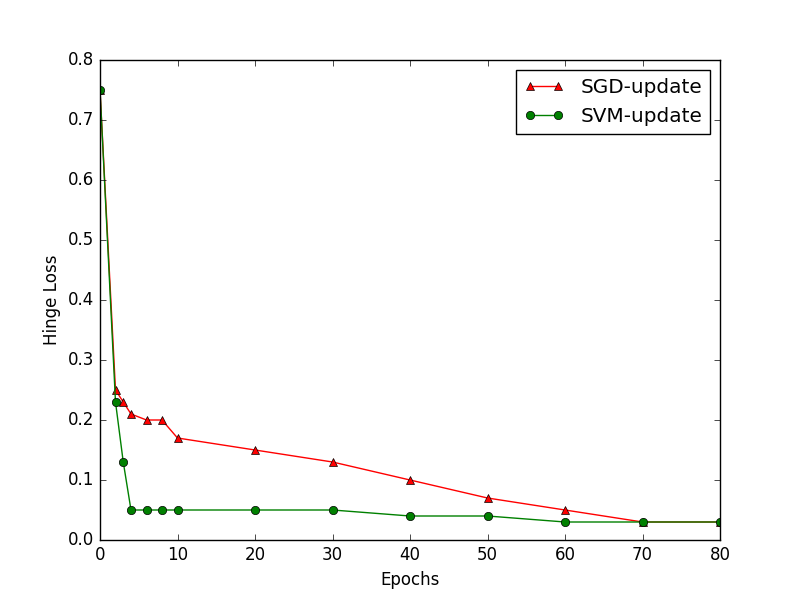}
    \caption{\textbf{Left}: mnist \& \textbf{Right}: cifar. We can see the hinge loss over train set versus epochs. We observe that our svm-based updates converge much faster than the traditional back-propagation using sgd. 
    % We should mention that in the case of mnist we converge two orders of magnitude faster than sgd in terms of epochs, and in the case of cifar we converge one order of magnitude faster.
    }
    \label{fig:hingeloss}
\end{figure}
\begin{figure}[]
    \centering
    \includegraphics[scale=0.30]{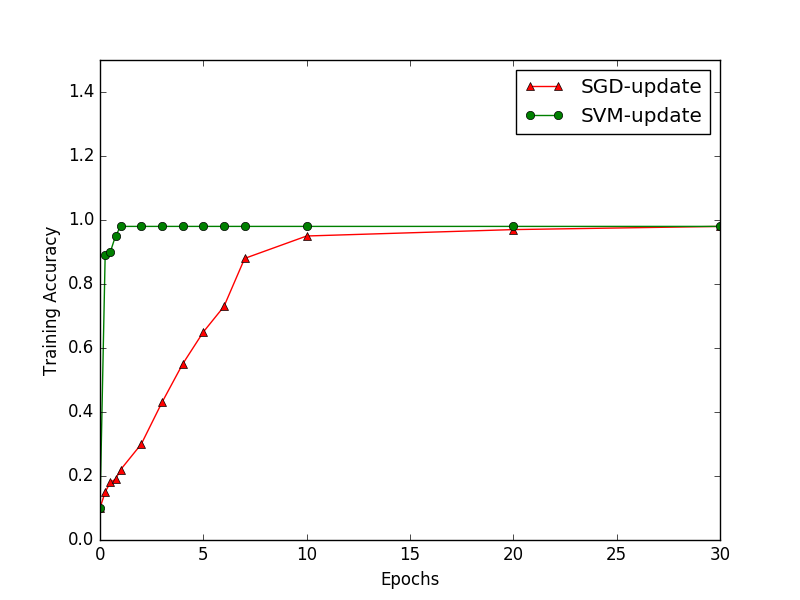}
    \includegraphics[scale=0.30]{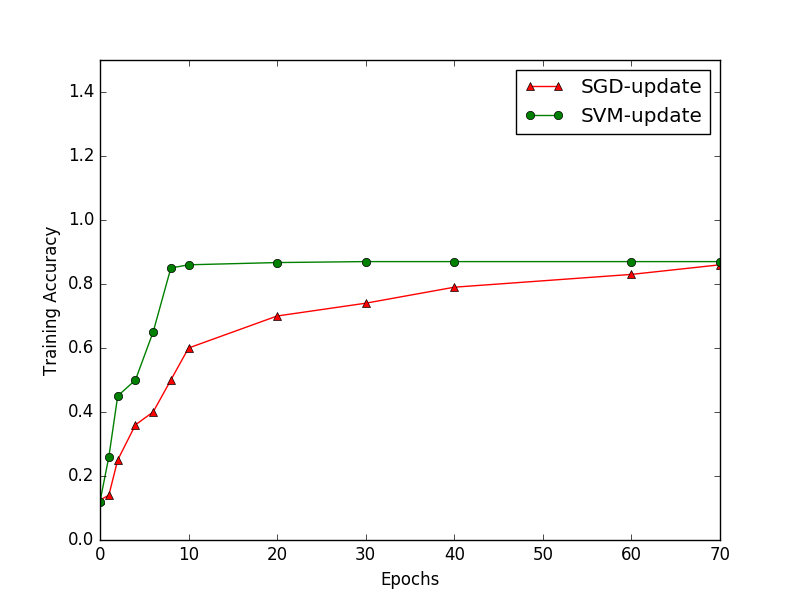}
    \caption{\textbf{Left}: mnist \& \textbf{Right}: cifar. We can see the training accuracy over the entire train set versus epochs.  We can see that our svm based updates converge much faster than the traditional back-propagation using sgd. 
    % We should mention that in the case of mnist we converge two orders of magnitudes faster than sgd in terms of epochs, and in the case of cifar we converge one order of magnitude faster.
    }
    \label{fig:accuracy}
\end{figure}

\subsection{Experimental Settings}
We used deep convolution networks for both mnist and cifar-10 classification task. For mnist, we constructed a network with two convolution layers of size 5x5 followed by a dense layer of size 30. The number of convolution filters used were 64 and 32 respectively. We experimented with different number of layers, and different layer sizes but these gave the best results. For cifar, we constructed a slightly deeper network with 3 convolution layers of size 5x5 with depth 3 followed by a dense layer of size 30. The number of convolution filters used in each convolution layer were 120, 60 and 30 respectively. In both the networks, all the hidden layers are followed by a ReLU non-linearity and maxpooling layer of size 2x2 with strides of 2x2.  

The hinge loss for a sample was computed as the average hinge loss over all output nodes.  We set aside a separate validation set for parameter tuning. The parameters pertaining to the best performance on validation set were saved for prediction. We had a test set that was kept unseen during the entire training and development process. We experimented with different types of activations apart from ReLU but did not observe a significant difference in the performance.

% \begin{figure*}
%     \centering
%     \includegraphics[scale=0.40]{images/train_acc_mnist.png}
%     \includegraphics[scale=0.40]{images/train_acc_cifar.png}
%     \caption{We can see the training accuracy over the entire train set versus epochs for mnist (left) and cifar-10 (right) datasets. We can observe that our svm based updates converge much more faster than the traditional back-propagation using adams optimizer. We should mention that in the case of mnist we converge two orders of magnitudes faster than sgd in terms of epochs, and in the case of cifar-10 we converge one order of magnitude faster.}
%     \label{fig:accuracy}
% \end{figure*}

\subsection{Results}
We showed in Sec \ref{sec:gen} that we can compute generalized updates from the minibatch that lead to decrease in the loss over entire training set, as well as test set. 
% We can see in Figure \ref{fig:ratio} that the updates are indeed generalized, which vindicates our theoretical assertions. It shows that the ratio of decrease in train set loss to the minibatch loss is $\approx$ 1. 
% We should note that we obtain the updates by solving an svm, which tries to minimize the $l_2$ norm of the update that serves the dual purpose of getting a generalized update, plus keeping the update small enough to not trigger the second order effects in the deep network. We should mention that towards the end of optimization the algorithm requires greater number of samples in the minibatch in order to obtain generalization. This gives a theoretically sound way of obtaining the minibatch sizes using Theorem 2. 
% The biggest achievement of the proposed method is in terms of the speed of convergence. 
We can see in Figure \ref{fig:hingeloss} that the loss decreases much faster using our algorithm as compared to sgd. We observed a speed up of nearly two order of magnitude in terms of epochs required for convergence in mnist dataset, and one order of magnitude in convergence over cifar dataset. We can see in Figure \ref{fig:accuracy} that the same trend follows for training accuracy. 

% We observe that we achieve two orders of magnitude speed up for mnist and one order of magnitude speed up for cifar. 
We require one epoch for convergence over mnist as compared to thirty epochs required by sgd optimizer. Similarly, we require five epochs for convergence over cifar as opposed to seventy epochs required by sgd optimizer. These are significant speed ups, and even though our updates are more expensive, they can still give a significant overall speed up over traditional back-propagation. 

We do not observe  significant differences in the prediction accuracy of both sgd and our approach over test set. Both methods lead to similar accuracy over the test set as can be seen in Table \ref{tab:acc}. We used the same validation set to tune the parameters for both the methods. But, as can be seen in Figure \ref{fig:genandval}, we do not require early stopping with our algorithm, which is another advantage of our method. As our updates are generalized we do not need to do nested validation in order to obtain the best results. We only show the results over mnist due to space constraints but results on cifar were similar.

\begin{table}[]
    \centering
    \begin{tabular}{c|c|c}
         Dataset & svm-update & sgd-update \\ \hline
         mnist & 0.989 & 0.982\\ \hline
         cifar & 0.878 & 0.867\\ 
    \end{tabular}
   
    \caption{We can see that the results obtained on the test set are only slightly different. Both the methods were tuned over a validation set of same size, and the best performing parameters were chosen to evaluate over test set. \textbf{Note} that we are not using the state-of-the-art networks for the tasks but standard deep convolutional networks.}
    \label{tab:acc}
\end{table}

\subsection{Robustness}
One very important aspect of a learned network in deep learning is robustness to noise. We observed that both the update algorithms i.e. traditional back propagation and our svm-based updates perform well under random noise and do not show divergent behaviour. But, we observed that the proposed svm-update algorithm made the learned network more robust to additive adversarial noise. Robustness to adversarial noise is important to protect the network during an attack. We can see in Table \ref{tab:noise} that the norm of the noise required to force the network into misclassification is much higher in our case compared to sgd. It shows that our generalized updates lead to a more robust network.

\begin{table}[]
    \centering
    \begin{tabular}{c|c|c}
         Norm & sgd-based & svm-based  \\\hline
         frobenius &  5.12 & 7.32 \\ \hline
         infinity & 5.90 & 8.11 \\ \hline
         nuclear & 7.29 & 8.11 \\ \hline
         1-norm &  6.23 & 7.59 
    \end{tabular}
    \caption{We give details of the additive adversarial noise learned for mnist using traditional back-propagation and svm-based updates. Additive adversarial noise is the minimum amount of noise to be added to images such that the network misclassifies them. 
    % It is learned by keeping the network fixed and learning over regularized additive noise to obtain perfect misclassification. 
    }
    \label{tab:noise}
\end{table}

\subsection{Data-dependence}
\begin{figure}[]
    \centering
    \includegraphics[scale=0.30]{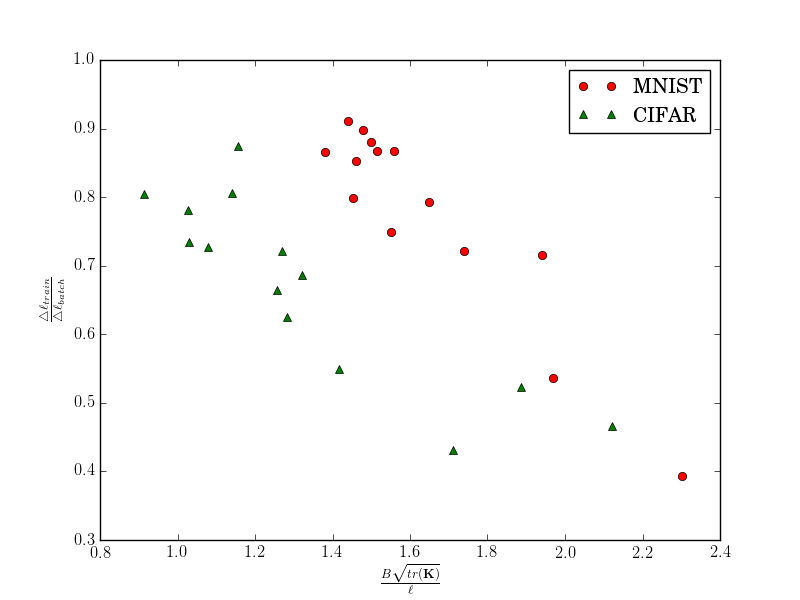}
    \includegraphics[scale=0.30]{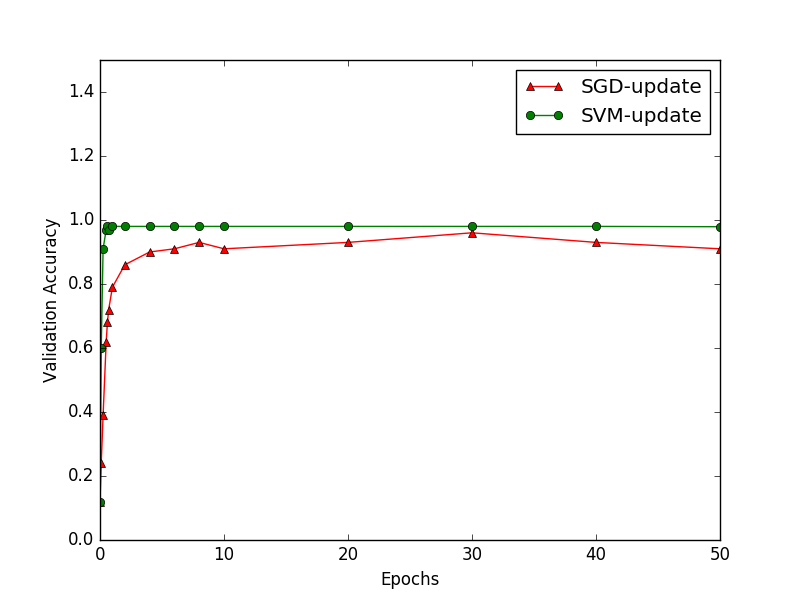}
    \caption{\textbf{Left}: We see that the ratio of decrease in loss over train set and mini-batch decreases with increase in the bound (see Theorem \ref{margthm}), implying that the updates become less generalized. \textbf{Right}: We observe that the validation accuracy initially increases and then  stabilizes for mnist using our algorithm, as opposed to sgd.}
    \label{fig:genandval}
\end{figure}

We experiment with fitting random labels to a network using sgd and our proposed svm-updates. It is well known that deep neural networks have high finite sample expressivity \cite{zhang2016understanding} and can memorize even randomly assigned labels \cite{zhang2016understanding}. We show in Section \ref{sec:gen} that our updates are generalized and the bounds depend on the data. Despite that we decided to conduct experiments to observe the behaviour of our algorithm when faced with randomly assigned labels. We hoped that the bounds generated for a minibatch in such a scenario would grow too large and indicate that the data can not be used for learning the network. We indeed observed that sgd was able to overfit the randomly assigned labels but our bound grew larger. We can see in Figure \ref{fig:genandval} that bound grows larger when the loss is only decreasing over the minibatch, as opposed to decreasing over the entire training set.

\section{Conclusion}
We develop an algorithm based on generalized updates that can lead to faster convergence in deep networks. We prove that the updates are guaranteed to decrease the loss over the train and test set under certain conditions. Specifically, we achieve two orders of magnitude speed up compared to back-propagation over the same network on one of the dataset and one order of magnitude speed up over the other. We also show that the learned network is more robust to adversarial noise and over-fitting. We provide theoretical bound on the error of the network after $k$ updates. In the future, we hope to extend the idea towards speeding up convergence in RNNs that are far more slower to train compared to deep CNNs.

\section{Acknowledgement}
The Authors acknowledge support of the UK Defence Science and Technology Laboratory (Dstl) and Engineering and Physical Research Council (EPSRC) under grant EP/R018693/1, Semantic Information Pursuit for Multimodal Data Analysis.  This is part of the collaboration between US DOD, UK MOD and UK EPSRC under the Multidisciplinary University Research Initiative.
\bibliographystyle{abbrv}
{
\bibliography{main}
}

\end{document}